\documentclass[10pt,twocolumn,letterpaper]{article}

\usepackage{cvpr}
\usepackage{times}
\usepackage{epsfig}
\usepackage{graphicx}
\usepackage{amsmath}
\usepackage{amssymb}

\usepackage{cite}
\usepackage{times}
\usepackage{epsfig}
\usepackage{graphicx}
\usepackage{algorithmic}
\usepackage{amsmath,amssymb,amsfonts}
\usepackage{enumitem}
\usepackage{url}
\usepackage{booktabs}

\usepackage{subfigure}
\usepackage{multirow}
\usepackage{ifthen}
\usepackage{ulem}
\usepackage{makecell}
\usepackage[utf8]{inputenc}
\usepackage{multirow}

\usepackage[pagebackref=true,breaklinks=true,letterpaper=true,colorlinks,bookmarks=false]{hyperref}

 \cvprfinalcopy 

\def\cvprPaperID{****} 

\ifcvprfinal\fi
\begin{document}

\title{Do We Need Fully Connected Output Layers in Convolutional Networks?}

\author{Zhongchao Qian$^{1}$ \quad Tyler L. Hayes$^{1}$ \quad Kushal Kafle$^4$ \quad  Christopher Kanan$^{1,2,3}$\\
$^1$Rochester Institute of Technology \quad $^2$Paige \quad $^3$Cornell Tech \quad $^4$Adobe Research\\
}

\maketitle

\begin{abstract}

Traditionally, deep convolutional neural networks consist of a series of convolutional and pooling layers followed by one or more fully connected (FC) layers to perform the final classification. While this design has been successful, for datasets with a large number of categories, the fully connected layers often account for a large percentage of the network's parameters. For applications with memory constraints, such as mobile devices and embedded platforms, this is not ideal. Recently, a family of architectures that involve replacing the learned fully connected output layer with a fixed layer has been proposed as a way to achieve better efficiency. In this paper we examine this idea further and demonstrate that fixed classifiers offer no additional benefit compared to simply removing the output layer along with its parameters. We further demonstrate that the typical approach of having a fully connected final output layer is inefficient in terms of parameter count. We are able to achieve comparable performance to a traditionally learned fully connected classification output layer on the ImageNet-1K, CIFAR-100, Stanford Cars-196, and Oxford Flowers-102 datasets, while not having a fully connected output layer at all.

\end{abstract}

\section{Introduction}

\begin{figure}[t]
\begin{center}
\includegraphics[width=0.9\linewidth]{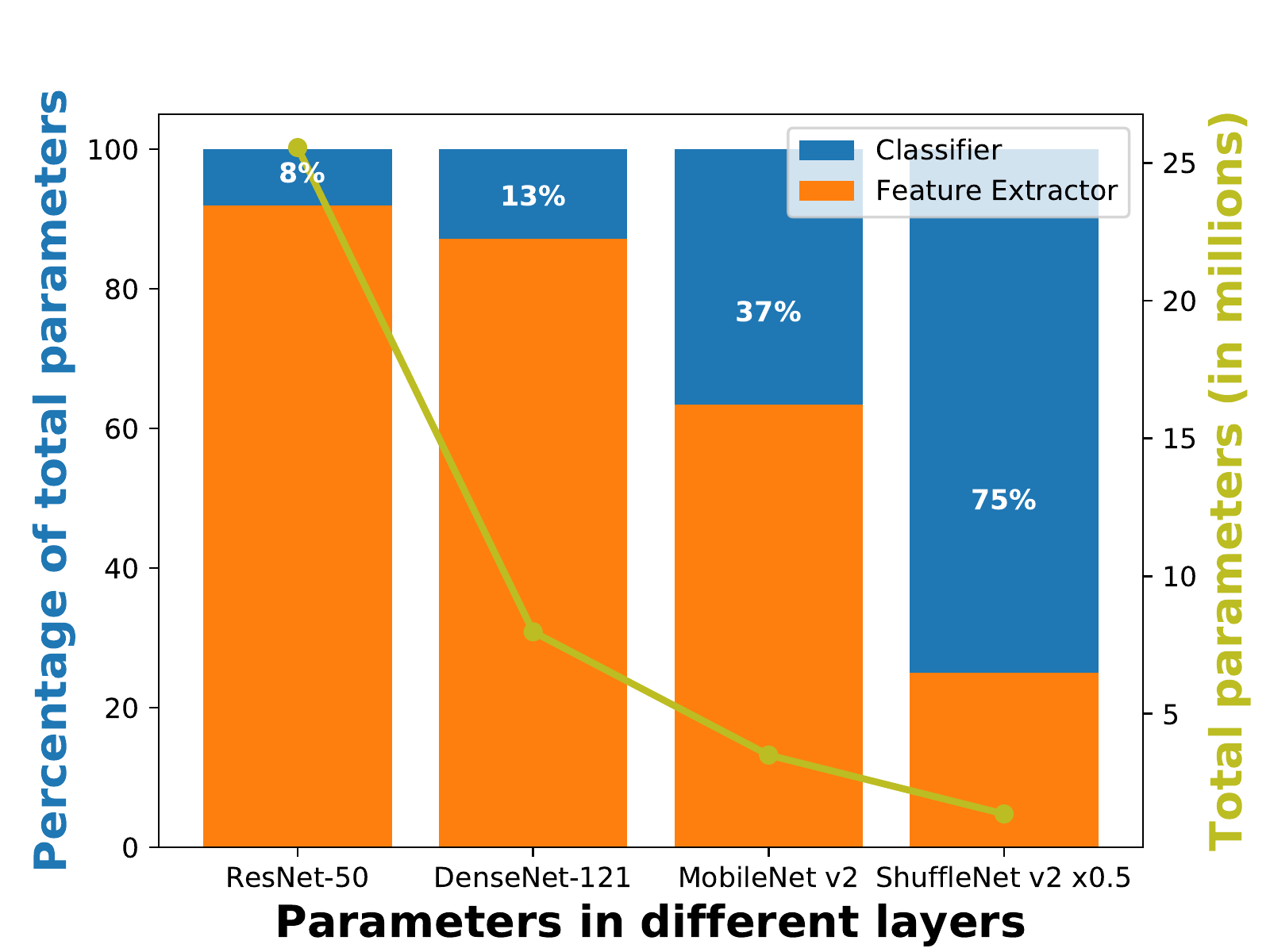}
\end{center}
   \caption{Bar plot showing the percentage of parameters in different parts of different architectures for ImageNet-1K classification. The green plot shows the total number of parameters for each architecture. As models get more efficient and compact, the final classifier accounts for more of the total parameters. Our method eliminates the need for a final fully connected (FC) layer for classification, significantly reducing our memory requirements, especially in already efficient models.
}
\label{fig:parameters-per-architecture}
\end{figure}

The strong performance of deep convolutional neural networks (CNNs) has enabled an enormous number of new computer vision applications. However, many state-of-the-art CNN architectures are ill-suited for deployment on mobile and embedded devices due to their high computational and memory requirements. The vast majority of CNN architectures are designed as having a feature extractor followed by a classifier. The feature extractor consists of convolutional layers and pooling operations, while the classifier is made up of one or more fully connected layers. A number of papers have developed methods for reducing the parameters in the feature extractor, for instance group convolutions first implemented in AlexNet~\cite{krizhevsky2012imagenet}, depth-wise separable convolutions introduced in Xception~\cite{chollet2017xception}, and squeeze and expand operations from SqueezeNet~\cite{iandola2016squeezenet}, but little work has been done to reduce the parameters in the classifier's fully connected layers. Because the number of parameters in the classifier are typically proportional to the number of categories, the classifier can consume a large portion of the network's total parameters for large datasets. For example, in MobileNet-v2 the fully connected layers consume 37\% of the parameters in the CNN for ImageNet-1K. For supervised lifelong machine learning applications~\cite{hayes2019memory,hayes2019lifelong,parisi2019continual,kemker2017fearnet}, the number of categories increases over time so having the number of parameters increase sub-linearly with the number of classes learned is especially important.

There has been recent interest in fixing the classifier weight matrix~\cite{hoffer2018fix, pernici2019fix}. These methods initialize the weights, but do not update them during training, thus increasing the efficiency of models. In this paper, we take the idea further. We use a fixed identity matrix as the classifier, which is equivalent to removing the classifier layer rather than having a feature extractor followed by a classifier. We directly train the convolutional layers for classification and entirely eliminate the traditional classification layer. We show that the number of parameters can be greatly reduced by rethinking the design (Fig.~\ref{fig:parameters-per-architecture}).

\textbf{Our main contributions are:}
\begin{enumerate}[noitemsep, nolistsep]
    \item We show that the final convolutional layer can be modified in many widely used CNN architectures to enable the fully connected layer to be completely eliminated, with little loss in classification performance but with a large reduction in the total number of parameters for many-class datasets. 
    
    \item We compare our method against existing fixed classifier methods and achieve superior results, while being much simpler and more efficient.
    
    \item We show that the final classifier layer contributes little to overall model classification accuracy. We propose that using a fully connected layer is very inefficient and should be changed in future architecture designs for image classification.

\end{enumerate}

\section{Related Work}

Our work relates to two main categories of existing work: \textbf{1) Alternative classifiers} which have been explored mainly for the purposes of making the output layer fixed and/or making the classifier more discriminative and \textbf{2) Parameter reduction techniques} which range from ground-up redesign of networks to post-trained pruning techniques. We discuss these categories of related work in detail in the following sections.
\subsection{Alternative Classifiers}

In \cite{springenberg2014striving}, a study was conducted to understand what components of a CNN are absolutely necessary. They concluded that a CNN can be constructed using only convolution operations by demonstrating that the final fully connected output layer could be replaced by 1-by-1 point-wise convolutions; however, they did not consider that the entire classification layer could be removed.

A few existing works have studied how to reduce the number of parameters in a CNN's classifier for many-class datasets by using fixed output matrices~\cite{hoffer2018fix,pernici2019fix}. In \cite{hoffer2018fix}, it was shown that any fixed orthogonal output matrix could be used to replace a learned output matrix with no reduction in performance. While this does not reduce the number of parameters or computational requirements, they then demonstrated that a Hadamard matrix could be used, enabling increased efficiency. However, it is not possible to construct a Hadamard matrix if the input to the classifier has fewer dimensions than the number of output categories because a Hadamard matrix's rows and columns are mutually orthogonal. This means for ResNet-18, which has 512 dimensional features input to the classifier, it would be limited to classifying at most 512 categories. This limitation was overcome in \cite{pernici2019fix}, which proposed a different method of creating a fixed output classifier. Their approach uses coordinate values of high-dimensional regular polytopes as rows of the fixed classifier weight matrix. While this approach works, it can be difficult to train, and it is used mainly to optimize for feature extraction.

It is not currently clear which fixed output matrix approach is best, and some of these methods still require the classifier's parameters to be stored, even if the parameters are not updated during training. In contrast, our approach avoids using an explicit classification layer entirely, eliminating the problem of selecting and storing a fixed classifier weight matrix.

\subsection{Parameter Reduction Techniques}

A popular method for reducing the number of parameters in the feature extractor is by using variants of convolution. Group convolutions split the convolution input and output channels into groups, where each group is a convolution operation independent of other groups~\cite{krizhevsky2012imagenet}. By removing connections between channels belonging in different groups, it reduces parameters in the convolution by a factor equal to the number of channels. Depth-wise separable convolution is a two step procedure. First, there is a group convolution where the number of input channels, output channels, and groups are all the same, followed by a point-wise convolution with the desired number of output channels~\cite{chollet2017xception}.

Other methods for reducing the number of parameters are pruning and quantization. Pruning removes (zeros out) weights after training to promote sparsity, and a wide variety of pruning methods have been explored~\cite{lecun1990optimal,han2015learning,louizos2018learning,srinivas2015data,alvarez2016learning,hu2016network,li2017pruning}. Quantization methods typically reduce the numeric precision of the weights after training, which can greatly reduce the number of parameters~\cite{courbariaux2016binarized,kim2016bitwise,jacob2018quantization}. Both pruning and quantization are complementary to our method, which focuses on eliminating the classifier to reduce the number of parameters.

\section{Method}
\label{sec:method}

We evaluate three different fixed classifiers: 1. using a fixed orthogonal projection; 2. using a fixed Hadamard projection; and 3. removing the fully connected layer, which is equivalent to using a fixed identity matrix for projection and setting the bias term to zero. We briefly describe the first two methods~\cite{hoffer2018fix}. We then describe our implementation. We compare all three fixed classifiers against a learned fully connected classifier, and against each other, to evaluate their effects on the model.
 
\begin{figure*}[ht]
\centering
\includegraphics[width=0.95\linewidth]{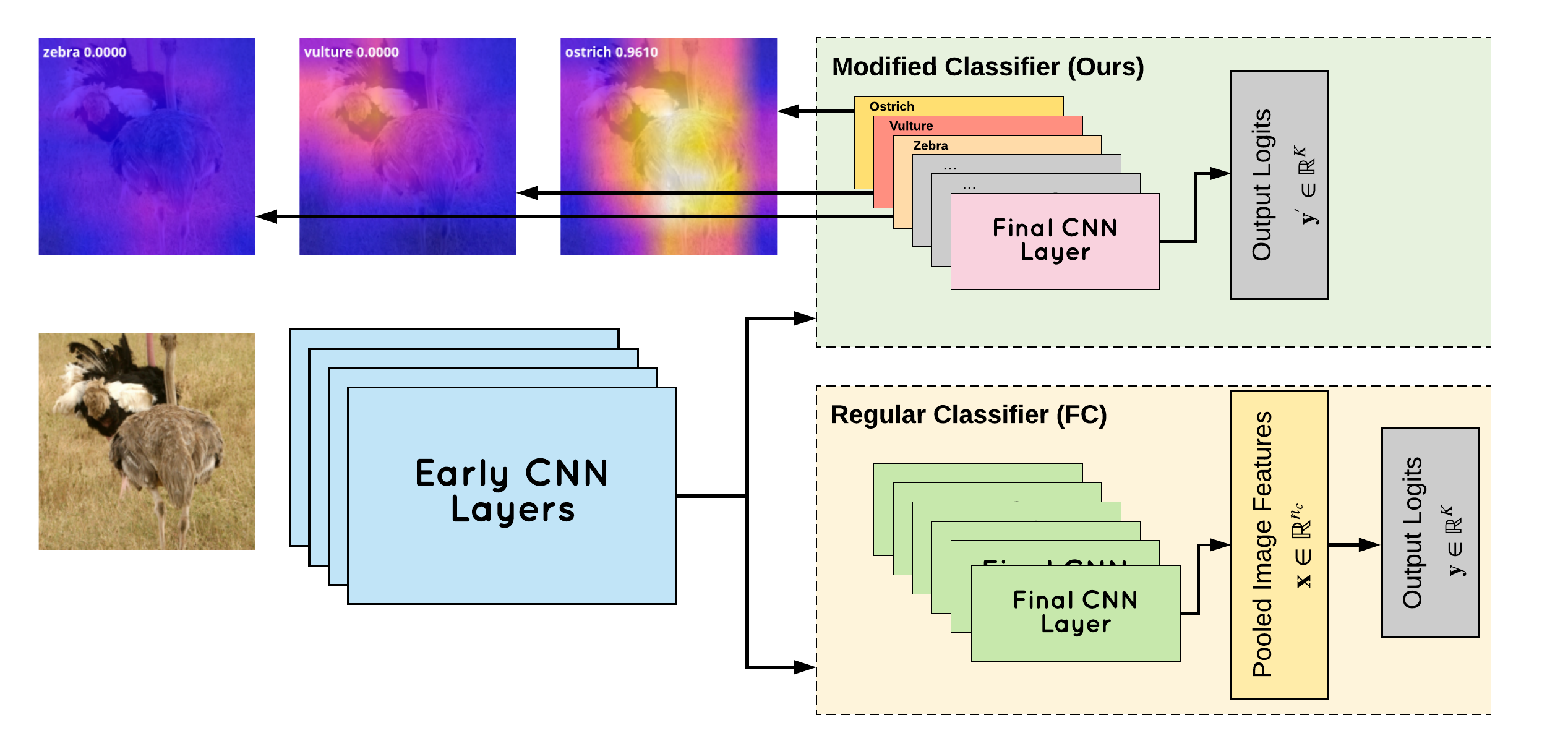}
   \caption{A depiction of our method. We perform global average pooling on the final CNN layer to obtain scores for classification without using a FC layer. This reduces the number of parameters in the network. Moreover, since each channel in the output of the final CNN layer represents an output class, they can be directly visualized to represent class-specific visualizations. We obtain high activation for regions with the correct class (ostrich), low activation for an unrelated class (zebra), and regions containing background objects (vulture). 
   }
\label{fig:our-method}
\end{figure*}

\subsection{Learned Fully Connected Classifier}

In typical deep neural networks for single-class image classification, the last layer is a fully connected layer of affine transformation, and all its parameters are learned. The input vector to the classifier, $\mathbf{x}$ is multiplied with the weight matrix $\mathbf{W}$, then the bias vector $\mathbf{b}$ is added to the result, to produce the output vector $\mathbf{y}$:
\begin{equation}
    \mathbf{y}=\mathbf{W}\mathbf{x}+\mathbf{b} \enspace .
\end{equation}
Typically, $\mathbf{x}$ could be the result of a previous fully connected layer, or more commonly nowadays, the results of global average pooling of the feature maps obtained by previous convolution layers. The weight matrix $\mathbf{W}$ and bias $\mathbf{b}$ are optimized during back-propagation using gradient descent. The output vector $\mathbf{y}$ usually goes through softmax activation to obtain the classification likelihood for each potential category.

\subsection{Fixed Orthogonal Classifier}

In a fixed orthogonal classifier~\cite{hoffer2018fix}, everything is the same as using a learned fully connected classifier, except for the weight matrix $\mathbf{W}$, which is initialized using a random orthogonal matrix. This orthogonal matrix is obtained through QR decomposition of a random real square matrix. During back-propagation, it is not updated, hence the name fixed orthogonal classifier. In the case where the input and output sizes differ, $\mathbf{W}$ is a semi-orthogonal matrix instead.

\subsection{Fixed Hadamard Classifier}
In fixed Hadamard classifiers~\cite{hoffer2018fix}, the weight matrix is also fixed (i.e., not updated), and it is initialized as a Hadamard matrix. In this case, the Hadamard matrix is constructed using Sylvester's construction. Let $\mathbf{H}_1$ be a Hadamard matrix of order 1, defined as
\begin{equation}
    \mathbf{H}_1=\begin{bmatrix}
1
\end{bmatrix}.
\end{equation}
Let $k$ be any non-negative integer greater than 1. Higher order Hadamard matrices of order $2^k$ can be constructed using Hadamard matrices of the lower order $2^{k-1}$, given as,
\begin{equation}
\mathbf{H}_{2^k}=\begin{bmatrix}
\mathbf{H}_{2^{k-1}} & \mathbf{H}_{2^{k-1}} \\
\mathbf{H}_{2^{k-1}} & -\mathbf{H}_{2^{k-1}}
\end{bmatrix}.
\end{equation}
By iterating this process, we can obtain Hadamard matrices of order 1, 2, 4, \ldots, $2^k$.

To construct the weight matrix, we would need to obtain a Hadamard matrix of order $2^k$, where $k=\lceil \log_2{\max(n_c, K)} \rceil$. Here $n_c$ is the number of channels from the last convolution layer, also the dimension of the input $\mathbf{x}$, and $K$ is the number of classification categories, also the dimension of $\mathbf{y}$. Then the matrix is truncated to fit the size of the input and output, by taking its first $n_c$ rows and first $K$ columns.

Then the output is obtained using the following calculation:
\begin{equation}
\mathbf{y}=\alpha\mathbf{W}\mathbf{x}+\mathbf{b} \enspace ,
\end{equation}
where $\alpha$ is a learned scalar parameter that is updated during back-propagation.

The fixed Hadamard classifier using this construction has a limitation. It cannot produce effective outputs when the output dimension is larger than that of the input. For instance when using it in ResNet-18 for classification of ImageNet-1K, the input is a vector of 512 dimensions, while the output needs to be 1000 dimensions. Here $\mathbf{W}$ has 1000 rows and 512 columns, and it is apparent that rows 513 through 1000 are identical to rows 1 through 488, resulting in the same intermediate results for all these items. The final results only differ because $\mathbf{b}$ could be different.

\subsection{Fixed Identity Classifier}

In our method, we remove the final fully connected layer completely, and use the output from the global average pooling layer directly to compute classification scores. The global average pooling layer is immediately after the last convolution layer. A depiction of our method is shown in Fig.~\ref{fig:our-method}. Implementation wise, it is equivalent to setting the weight matrix $\mathbf{W}$ as an identity matrix $\mathbf{I}$, where all the elements on the diagonal are 1 and all other elements are 0. This matrix is not updated throughout training. We also drop the bias term, $\mathbf{b}$.
 
Our method offers an additional benefit: it enables the CNN outputs to be visualized immediately, similar to class activation mappings (CAM)~\cite{zhou2016learning}. Contrary to CAM, which requires post processing intermediate results from the neural network, we can obtain these visualizations without any extra compute, during the forward pass (inference), along with obtaining the classification scores. The visualization results are demonstrated in Figure~\ref{fig:our-method}, using an image of an ostrich from the ImageNet-1K test set.

Our method suffers the same limitation as a fixed Hadamard classifier: it is unable handle cases where the number of classification categories is greater than the number of channels from the last convolution layer. However, we are not promoting the method as a drop in replacement on existing architectures. Rather, it serves as a proxy tool to study the final classifier layer in current image classification architectures, and a possible way to design classifiers for future efficient architectures.

\section{Experiments}

\subsection{Architectures}

We evaluate our method on several common CNN architectures and datasets. We chose several common residual networks, as well as mobile architectures that contain far fewer parameters. We compare our method using the following CNN architectures and use their respective PyTorch implementations:
\begin{itemize}[noitemsep, nolistsep]
    \item \textbf{ResNet-18} -- The ResNet-18 architecture is a common residual network consisting of 18 layers and skip connections to help gradient flow~\cite{He_2016_CVPR}. We chose this network since it is the fastest residual network to train.
    \item \textbf{ResNet-50} -- ResNet-50 is a residual network with 50 layers and skip connections~\cite{He_2016_CVPR}. We chose this architecture since it has been commonly used for computer vision applications and achieves higher performance on ImageNet than ResNet-18.
    \item \textbf{ResNet-32} -- This variant of ResNet is one variant that is optimized for the CIFAR image classification dataset, where the input image size differs than that used in ResNet-18 and ResNet-50.
    \item \textbf{DenseNet} -- The Dense Convolutional Network takes the skip connection idea further~\cite{huang2017densely}. In DenseNets, each layer has a skip connection to every other layer in a feed forward fashion.
    \item \textbf{MobileNet-v2} -- MobileNet architectures are designed to efficiently run on mobile devices by replacing convolutional layers with depth-wise separable convolutions. We use the MobileNet-v2 architecture~\cite{sandler2018mobilenetv2}, which additionally uses bottlenecks and residual connections. We chose this architecture since it is computationally efficient and we further reduce the network's memory requirements with our method.
    \item \textbf{ShuffleNet-v2 x0.5} -- ShuffleNet architectures use point-wise group convolutions and bottleneck layers to run efficiently on mobile devices. A channel shuffle operation is applied on top of these operations to allow gradients to flow between different channel groups, which improves accuracy. ShuffleNet-v2 additionally introduces a channel split operation~\cite{ma2018shufflenet}. We use v2 with a half width (x0.5).
\end{itemize}

\subsection{Datasets}

We perform our main experiments on the ImageNet-1K dataset, demonstrating the robustness of our method on a large dataset with many categories. Additionally, we perform experiments on CIFAR-100 and also provide results on two smaller datasets to demonstrate our method's ability to perform transfer learning.
These datasets were chosen because they have a large number of classes, allowing us to test our method's capability of performing well, while also saving memory. We chose the following datasets:
\begin{itemize}[noitemsep, nolistsep]
    \item \textbf{ImageNet-1K} -- The ImageNet dataset consists of images from 1,000 categories from the internet~\cite{russakovsky2015imagenet}. Each category consists of 732-1,300 training examples and 50 validation examples, which are used for testing. This is a common large-scale image classification dataset that allows us to test the ability of our method to scale up and showcase its parameter savings.
    \item \textbf{CIFAR-100} -- The CIFAR-100 dataset~\cite{Krizhevsky09learningmultiple} contains 100 classes each containing 600 color images of size $32\times32$. For each class, there are 500 images for training and 100 for testing.
    \item \textbf{Stanford Cars-196} -- The Stanford Cars dataset consists of 196 car classes with 8,144 training and 8,041 testing images~\cite{KrauseStarkDengFei-Fei_3DRR2013}.
    \item \textbf{Flowers-102} -- The Oxford Flowers dataset consists of 102 flower categories, with each class containing 40-258 images~\cite{nilsback2008flowers}.
\end{itemize}
While CIFAR allows us to quickly evaluate different methods, ImageNet tests the ability of our method to scale up to a large number of categories. The Stanford Cars-196 and Flowers-102 datasets allow us to test our method on fine-grained transfer learning tasks.

\subsection{Implementation Details}

\begin{table}[t]
\caption{Transfer learning parameter settings for each architecture.}
\label{tab:imagenet-params}
\begin{center}
\setlength\tabcolsep{3pt}
\begin{tabular}{lccc}
\toprule
& \textsc{Learning} & \textsc{Weight} & \textsc{Batch} \\
\textsc{Architecture}  & \textsc{Rate} & \textsc{Decay} & \textsc{Size} \\
\midrule
ResNet-18 & 0.01 & 1e-3 & 64 \\
ResNet-50 & 0.01 & 1e-4 & 64 \\
MobileNet v2 & 0.01 & 1e-4 & 64 \\
MNASNet x1.0 & 0.001 & 1e-4 & 32 \\
ShuffleNet v2 x0.5 & 0.1 & 1e-4 & 64 \\
\bottomrule
\end{tabular}
\end{center}
\end{table}

\begin{table*}[t]
\caption{Results on CIFAR with different models and different types of classifiers.}
\label{table:cifar-results}
\begin{center}
    \begin{tabular}{cllcc}
        \toprule
         \textsc{$K$} & 
         \textsc{Architecture} & \textsc{Classifier} &
         \textsc{Top-1 Accuracy} & \textsc{Performance Gap}  \\
         \midrule
         
         \multirow{6}{*}{100} 
         & \multirow{2}{*}{ResNet-32} 
         & Learned & 69.46\% & N/A \\
         & & Fixed Orthogonal & 68.61\% & -0.85\% \\\cline{2-5}
         
         & \multirow{4}{*}{DenseNet-BC} 
         & Learned & 77.61\% & N/A \\
         & & Fixed Orthogonal & 76.68\% & -0.93\% \\
         & & Fixed Hadamard & 75.84\% & -1.77\% \\
         & & Fixed Identity & 76.90\% & \textbf{-0.71\%} \\
         \midrule
         
         \multirow{4}{*}{64}
         & \multirow{4}{*}{ResNet-32} 
         & Learned & 73.94\% & N/A \\
         & & Fixed Orthogonal & 73.92\% & -0.02\% \\
         & & Fixed Hadamard & 73.97\% & +0.03\% \\
         & & Fixed Identity & 74.25\% & \textbf{+0.31\%} \\

         \bottomrule
    \end{tabular}
\end{center}
\end{table*}

We use PyTorch for all of our experiments. For CIFAR-100, every model on every architecture is trained from scratch. For the ImageNet results using a standard fully connected classification layer, we report the numbers from the PyTorch pre-trained models. For other classifiers on ImageNet, the models are trained from scratch. For all other experiments, we first initialize each model with pre-trained ImageNet weights and then fine-tune the network on the target dataset.

For training on ImageNet and CIFAR-100, we follow the original setup including methods for data augmentation~\cite{He_2016_CVPR, huang2017densely, hoffer2018fix}. For instance, for training ResNet-32 and DenseNet-BC on CIFAR-100, we do the following data augmentation for training: 4 pixels are padded on each side, then a mirroring is applied at random, followed by cropping to $32\times32$ randomly. For testing, the original image is used and only normalization is applied. This follows the practice in the respective works. For ResNet, the model is trained for 164 epochs, starting with a learning rate of 0.1, lowering it by a factor of 10 at 81 and 122 epochs. For DenseNet, it is trained similarly but for 300 epochs, lowering the learning rate after 150 epochs and 225 epochs. For other details, please refer to the respective original works. We provide parameters for the transfer learning experiments on Cars-196 and Flowers-102 in Table~\ref{tab:imagenet-params}.

\section{Results}
We evaluate all variants of the final classifiers on multiple architectures and multiple datasets, to compare and demonstrate their ability to perform image classification. We use the learned classifier as the baseline and measure the top-1 accuracy gap between the baseline, to see how much accuracy each classifier is sacrificing.

\subsection{CIFAR-100}

We trained models for different combinations of architectures and classifiers, and the results are shown in Table~\ref{table:cifar-results}. We train models on DenseNet for classification on CIFAR-100, with all variants of the classifier. We also trained ResNet-32 on CIFAR-100 using the learned classifier and the fixed Hadamard classifier, then we trained ResNet-32 with all variants of classifiers on 64 categories of the CIFAR-100 dataset. We follow the training methods from the original works.

We were unable to reproduce results for DenseNet-BC using the fixed Hadamard classifier, using the original open source code. In their original work, they report 77.67\% for the test accuracy, while we were only able to achieve 75.84\%. However, we believe our training setup is fair to all classifiers, therefore the performance gap still shows that not having a dedicated output layer is slightly superior to using a fixed Hadamard matrix, but not as good as having a learned fully connected classifier.

\subsection{ImageNet-1K}

\begin{table*}[t]
\caption{Results on ResNet-18 with each type of classifier, performing classification on ImageNet-1K and its subset.}
\label{table:rn18-1k-512}
\begin{center}
    \begin{tabular}{clcc}
        \toprule
         $K$ & \textsc{Classifier} & \textsc{Top-1 Accuracy} & \textsc{Performance Gap}  \\
         \midrule
         
         \multirow{2}{*}{1000} 
         & Learned & 69.76\% & N/A \\
         & Fixed Orthogonal & 66.48\% & -3.27\% \\
         \midrule
         
         \multirow{4}{*}{512}
         & Learned & 77.87\% & N/A \\
         & Fixed Orthogonal & 77.29\% & -0.58\% \\
         & Fixed Hadamard & 76.33\% & -1.53\% \\
         & Fixed Identity & 77.59\% & \textbf{-0.28\%} \\
         
         \bottomrule
    \end{tabular}
\end{center}
\end{table*}

We further move to a more challenging dataset by training and evaluating on ImageNet-1K. Here we use ResNet-18. Similar to the situation before, due to limitations of fixed Hadamard classifiers and our no classifier method (fixed identity classifier), we evaluate the full 1000 categories only on the learned classifier and the fixed orthogonal classifier. Then we evaluate all classifiers on the first 512 categories of ImageNet-1K, so that we can compare the Hadamard classifier and our fixed identity classifier. We follow the training method used in the ResNet paper, which is equivalent to training for 90 epochs with a batch size of 256. The results are shown in Table~\ref{table:rn18-1k-512}. The results indicate that while all fixed weights perform worse than learned weights, using a fixed identity matrix, which is equivalent to removing the classifier layer, outperforms both fixed orthogonal classifiers and fixed Hadamard classifiers.

\begin{table}[t]
\caption{Comparison of classification accuracy of the original ShuffleNet v2 and MobileNet v2 architectures with our method applied, trained on both the full dataset and 100 categories subset of ImageNet-1K.}
\label{table:imagenet-100-shufflenet}
\begin{center}
\resizebox{\linewidth}{!}{
\begin{tabular}{cllc}
    \toprule
    $K$ & \textsc{Architecture} & \textsc{Classifier} & \textsc{Top-1 Acc.} \\
    \midrule
    \multirow{4}{*}{1000} & \multirow{2}{*}{ShuffleNet V2 x0.5} & Learned & 60.55\% \\
    & & Fixed Identity & 53.06\% \\ \cmidrule{2-4}
    & \multirow{2}{*}{MobileNet v2} & Learned & 71.88\%\\
    & & Fixed Identity & 71.03\%\\
    \midrule
    
    \multirow{2}{*}{100}
     & \multirow{2}{*}{ShuffleNet V2 x0.5} & Learned & 72.94\%\\
     & & Fixed Identity & 74.42\%\\
     \bottomrule
\end{tabular}
}
\end{center}
\end{table}

Then we use MobileNet-v2 and ShuffleNet-v2 architectures, along with our fixed identity classifier on ImageNet-1K. In these two architectures, the final output layer accounts for most of the parameters. By removing the final layer, the model will see significant parameter savings. The results are shown in Table~\ref{table:imagenet-100-shufflenet}. We notice that there is a non-trivial degradation in performance. To evaluate whether this is due to the lack of parameters, or the modification to the architecture, we ran the same test on a very small subset of ImageNet, consisting of only 100 categories. We find that our fixed identity classifier does not perform worse than a learned classifier in this case, therefore the major performance gap on ImageNet-1K is likely due to the model being too small rather than the difference in the model architecture.
 
\subsection{Fine-Tuning with More Datasets}
\begin{table*}[t]
    \caption{Performance evaluation in terms of top-1 accuracy on Stanford Cars-196, and Flowers-102 for a standard classifier with a fully connected layer (Learned) and our modified classifier (Fixed Identity). We also report the parameter savings in terms of percentages for our method. For Cars-196 and Flowers-102, each result is the average of three runs.
    }
    \label{table:more-results}
    \begin{center}
    \begin{tabular}{lcccccc}
    \toprule
    &  \multicolumn{3}{c}{\textsc{Stanford Cars-196}}  & \multicolumn{3}{c}{\textsc{Flowers-102}}
    \\ 
    \cmidrule(r){2-7}
    & \textsc{Learned} & \textsc{Fixed Identity} & \textsc{Savings} & \textsc{Learned} & \textsc{Fixed Identity} & \textsc{Savings} \\
    \midrule
    ResNet-18 & 88.12\% & 86.06\% & 12.92\% & 93.42\% & 92.78\% & 16.83\% \\
    ResNet-50 & 89.90\% & 90.35\% & 5.66\% & 95.06\% & 94.64\% & 5.10\% \\
    MobileNet v2  & 87.68\% & 86.12\% & 24.26\% & 94.24\% & 93.95\% & 21.66\% \\
    ShuffleNet V2 x0.5 & 77.99\% & 75.76\% & 66.65\% & 87.75\% & 86.34\% & 63.52\% \\
    \bottomrule
    \end{tabular}
    \end{center}
\end{table*}

Finally, we demonstrate that our fixed identity classifier can be applied to more datasets and models. Results with several architectures on the Stanford Cars-196 and Flowers-102 datasets are shown in Table~\ref{table:more-results}. We show that our method works on ResNet-18, ResNet-50, MobileNet-v2, and ShuffleNet-v2 x0.5 on both datasets. The following results are obtained by fine-tuning a model pretrained on ImageNet. We can see that our method is able to achieve comparable results while using significantly fewer parameters, demonstrating its capabilities in transfer learning and generalization on more datasets.

\section{Discussion}

In this paper, we evaluated fixed classifier models which claim to be efficient in parameters and maintain performance. We compared fixed models and learned models. We then created a proxy model that constructs image classification neural networks by removing the fully connected layers from several modern CNN architectures and computed the classification scores directly from the final convolutional layer. This process is equivalent to having a fixed output layer that contains as little information as possible and is not updated with gradient descent. We used our model as a proxy tool to compare against models using a learned fully connected output layer and specifically designed fixed output layers.  Our results demonstrate that computing scores directly from the final convolutional layer performs better than using the Hadamard classifier. 

Using a fixed identity matrix greatly reduces the total number of parameters. For MobileNet and ShuffleNet, which already reduce the total number of parameters required by a model, we show that we can reduce these memory requirements even further (e.g., 39\% reduction for MobileNet-v2 and 75\% reduction for ShuffleNet-v2, both on ImageNet) with only a small degradation in performance, thus improving the efficiency of models.

We notice greater degradation of ImageNet-1K classification performance when using mobile architectures in conjunction with our method. In these scenarios, a significant percentage of parameters are removed from the model, and in the case of ShuffleNet-v2 x0.5, around 75\% parameters are removed, leaving the model with only 0.3M parameters, compared to 1.3M parameters of the vanilla model. Our results on ImageNet-100 showed that there is no performance degradation, which implies that the performance gap on ImageNet-1K is due to the model being too small to capture the statistics of the dataset. This suggests that while the final classifier layer uses a lot of parameters, it does not contribute much to the classification accuracy.

For our experiments, we computed classification predictions from the final convolutional layer by using global average pooling. However, more complex methods of pooling could be explored such as soft attention pooling~\cite{pawlowski2019needles}, which would allow the model to attend to more specific portions of a feature map to make final predictions. While soft attention pooling could possibly improve performance, it requires more parameters than global average pooling, making our approach less memory efficient. There are a number of alternatives we could explore to improve the performance of our method without increasing the number of parameters. One such way is to use orthogonal initialization of the final convolutional layer, which has shown similar convergence rates to unsupervised pre-training~\cite{xie2017all}. Similarly, orthogonal regularization could be used, which has been shown to improve network performance in terms of accuracy and stability of convergence~\cite{bansal2018can}. 

While we directly remove the fully connected layer of CNN architectures to improve our memory efficiency, we could additionally make use of network pruning~\cite{lecun1990optimal,han2015learning,louizos2018learning,srinivas2015data,alvarez2016learning,hu2016network,li2017pruning} to explicitly reduce parameters even further. Another option is to use network quantization to store parameters at a lower precision to save disk space and improve computational efficiency~\cite{courbariaux2016binarized,kim2016bitwise,jacob2018quantization}.

While our method yields comparable performance to a standard classifier when trained on ImageNet for all architectures tested, it comes with a caveat: it is incapable of handling more classes than the number of channels of output categories. However, this is not an issue, as fixed Hadamard classifiers also cannot do this, and the sole purpose of our model is to serve as a proxy tool to study the effectiveness of fixed classifier models, which claims to maintain performance while being more efficient.

Despite some caveats, our results suggest the final output layer does not need to be a learned fully connected layer.  We believe our results can be insightful for future efficient architecture design or neural architecture search. 

\section{Conclusion}

In this work, we evaluated the performance and efficiency of fixed classifier methods. We explored the elimination of the fully connected classifier with several modern CNN architectures. By using global average pooling to compute classification predictions directly from the final convolutional layer, we achieve comparable performance to several CNNs that use a fully connected layer, while greatly reducing the total number of parameters required by the model. This proves that specially designed fixed classifiers are not as effective as simply removing the final layer from networks, both in terms of parameter efficiency and classification accuracy. We showed that our approach is able to work on multiple datasets and neural network architectures. Finally, we demonstrated that the final classifier in general is not very efficient in terms of parameter size, and does not contribute very much to classification accuracy. We suggest future neural architecture designs should use output layers that are more efficient than fully connected layers.

\paragraph*{Acknowledgments.} This work was supported in part by the DARPA/MTO Lifelong Learning Machines program [W911NF-18-2-0263]. The views and conclusions in this work are those of the authors and should not be interpreted as representing the policies or endorsements of any sponsor.

{\small
\bibliographystyle{ieee_fullname}
\bibliography{library}
}

\end{document}